\documentclass{ecai}
\usepackage{graphicx}
\usepackage{latexsym}
\usepackage{times}
\usepackage{soul}
\usepackage{url}
\usepackage[hidelinks]{hyperref}
\usepackage[utf8]{inputenc}
\usepackage{graphicx}
\usepackage{amsmath}
\usepackage{booktabs}
\usepackage{algorithm}
\usepackage{algorithmic}
\usepackage{subfigure}
\usepackage{bm}
\usepackage{amssymb}

\hypersetup{hidelinks}

\begin{document}
\begin{frontmatter}
\title{Model-based Offline Policy Optimization with Adversarial Network}
\author[A,C]{\fnms{Junming}~\snm{Yang}}
\author[A,B]{\fnms{Xingguo}~\snm{Chen}}
\author[B]{\fnms{Shengyuan}~\snm{Wang}} 
\author[B]{\fnms{Bolei}~\snm{Zhang}~\thanks{Corresponding Author. Email: bolei.zhang@njupt.edu.cn}}

\address[A]{Jiangsu Key Laboratory of Big Data Security and Intelligent Processing, China}
\address[B]{School of Computer Science, Nanjing University of Posts and Telecommunications, China}
\address[C]{School of Modern Posts, Nanjing University of Posts and Telecommunications, China}

\begin{abstract}
Model-based offline reinforcement learning (RL), which builds a supervised transition model with logging dataset to avoid costly interactions with the online environment, has been a promising approach for offline policy optimization. As the discrepancy between the logging data and online environment may result in a distributional shift problem, many prior works have studied how to build robust transition models conservatively and estimate the model uncertainty accurately. However, the over-conservatism can limit the exploration of the agent, and the uncertainty estimates may be unreliable. In this work, we propose a novel \textbf{M}odel-based \textbf{O}ffline policy optimization framework with \textbf{A}dversarial \textbf{N}etwork (MOAN). The key idea is to use adversarial learning to build a transition model with better generalization, where an adversary is introduced to distinguish between in-distribution and out-of-distribution samples. Moreover, the adversary can naturally provide a quantification of the model's uncertainty with theoretical guarantees. Extensive experiments showed that our approach outperforms existing state-of-the-art baselines on widely studied offline RL benchmarks. It can also generate diverse in-distribution samples, and quantify the uncertainty more accurately.
\end{abstract}

\end{frontmatter}

\section{Introduction}
Over the last few years, reinforcement learning (RL) has achieved great success in a variety of simulation domains, e.g., Game of Go \cite{silver2018general}, Atari \cite{8490422}, MuJoCo \cite{salimans2017evolution}. However, vanilla RL methods have been struggling in many real-world applications, as they require frequent interactions with the environment, which are often costly or even dangerous. Offline RL provides an alternative approach that leverages logging datasets collected by another behavior policy for optimization without interacting with the online environment. With this innovative approach, RL can transcend its traditional boundaries and find application in an extensive array of real-world scenarios, becoming a practical and transformative solution in diverse fields like autonomous driving \cite{duan2020hierarchical}, recommendation \cite{zheng2018drn}, healthcare \cite{coronato2020reinforcement}, etc.

\begin{figure}
    \centering
    \includegraphics[width=8cm]{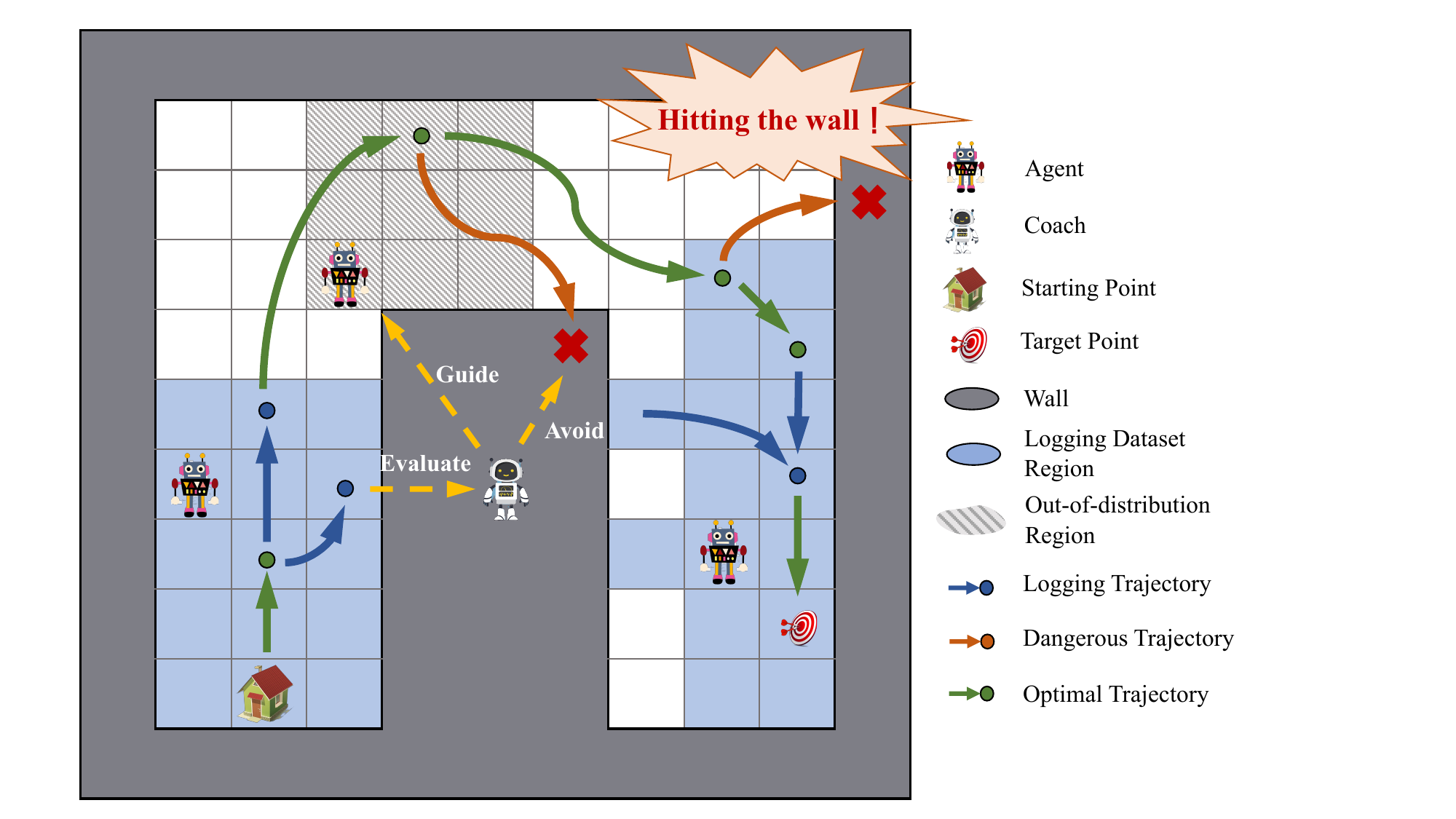}
    \caption{An agent optimizes its policy within fixed logging trajectories. Due to the limited scope of the dataset (blue), the agent may encounter new and unseen situations (white) during deployment, especially in out-of-distribution regions (grey slash). These situations can include obstacles such as hitting walls, leading to dangerous trajectories. The agent can have a higher success ratio to reach the target point if a coach can evaluate the agent's performance and guide the agent to optimize its trajectory.}
    \label{fig:intro}
\end{figure}

A straightforward approach for offline RL is adopting off-policy methods over the logging dataset directly. However, previous researches have demonstrated that the performance can be quite poor due to the \textit{distributional shift} problem \cite{levine2020offline,kumar2020conservative,wu2019behavior}, which occurs when the state-action distribution of the learning policy induced by the offline logging dataset differs from the distribution of the behavior policy. As a result, the off-policy algorithm cannot accurately evaluate the value function when selecting state-action pairs that are not covered in the offline dataset, as illustrated in Figure~\ref{fig:intro}.

To correct the distributional shift problem, the dominant paradigm is to introduce conservatism or uncertainty estimation in the offline RL algorithms. The related works can be broadly categorized as model-free offline RL and model-based offline RL. Model-free offline RL methods directly optimize the policy over the dataset and avoid the distributional shift problem by estimating conservative value functions \cite{wu2019behavior,kumar2020conservative,panaganti2022robust}. There may be 
a sample inefficiency problem as this conservatism can also limit generalization beyond the offline dataset.
Model-based approaches improve sample efficiency by first building a supervised transition model for interaction. The distributional shift problem can be mitigated by penalizing the reward with estimates of model uncertainty \cite{yu2020mopo,kidambi2020morel,wang2021offline,rigter2022rambo}. However, existing works typically rely on heuristic uncertainty estimation, which is unreliable for complex datasets. To train an offline policy that can be deployed to the online environment, it is crucial to generate diverse in-distribution samples and quantify the model uncertainty accurately at the same time. 
As depicted in Figure~\ref{fig:intro}, the agent has a higher success ratio to reach the target point if a coach can guide the agent to avoid the dangerous regions and evaluate the value function correctly.

In this paper, we propose a novel offline RL framework, named \textbf{M}odel-based \textbf{O}ffline policy optimization with \textbf{A}dversarial \textbf{N}etwork (MOAN). The basic idea is to introduce a two-player game in the model-based offline RL framework: one player is responsible for generating diverse transitions, and the adversary acts as a coach to accurately quantify the uncertainty. When the two players' policies converge, it is expected to generate diverse samples with accurate uncertainty quantification. In practice, the two-player game is implemented with a generative adversarial network. The discriminator learns to distinguish in-distribution data from out-of-distribution data, and the generator is trained to generate diverse samples that can confuse the discriminator. The output of the discriminator can be naturally implemented as uncertainty quantification for reward penalty. The lower bound of the policy performance in the real environment can be proved theoretically.

We conducted extensive experiments on the offline RL benchmark D4RL \cite{fu2020d4rl}. The results demonstrate that our proposed MOAN can achieve higher performance than state-of-the-art algorithms in most cases. In addition, our method has better generalization and can accurately quantify the model uncertainty.

The main contributions of this work are:
\begin{itemize}
    \vspace{-2mm}
    \item We propose a novel framework MOAN, to model the offline RL as a zero-sum two player game. The two players are expected to generate diverse in-distribution samples.
    \item Based on the proposed framework, we devise an accurate penalty to reshape the reward, with performance guarantee in the real online environment.
    \item Extensive experiments are conducted. The results show that our method outperforms state-of-the-art works in most cases.
    \vspace{-1mm}
\end{itemize}

\section{Related work}
The problem of offline RL has been extensively studied in the last few years. Prior works can be broadly categorized as model-free offline RL and model-based offline RL.
\vspace{-3mm}
\paragraph{Model-free offline RL} Model-free offline RL methods optimize the policy directly with the logging dataset. To overcome the distributional shift problem, several approaches have been proposed. One way is to use importance sampling to give the in-distribution samples higher importance, as in BCQ \cite{wen2020batch}, CQL \cite{kumar2020conservative}, AWR \cite{peng2019advantage} and VIP \cite{ma2022vip}. Alternatively, the target policy can be constrained to align with the behavior policy by adding a regularization term to the objective function that penalizes actions that deviate too much from the behavior policy \cite{schulman2015trust,wu2019behavior,kumar2019stabilizing,chen2023modified}, thereby providing a form of safe policy improvement guarantees. Some studies provide compelling empirical evidence for the benefits of these approaches \cite{fujimoto2021minimalist,kostrikov2021offline,shi2022pessimistic}. Based on these two directions, recent studies have also adopted ensemble models for a more robust value estimation \cite{derman2020bayesian,panaganti2022robust}. Adversarial training frameworks have also been used, where an adversary chooses the worst-case hypothesis (e.g., a value function or an MDP model) from a hypothesis class, and a policy player tries to maximize the adversarially chosen hypothesis \cite{cheng2022adversarially,xie2021bellman,bai2019model}. In model-free methods, there may be a data inefficiency problem, as the policy can only be optimized with the logging dataset.
\vspace{-3mm}
\paragraph{Model-based offline RL} Compared to model-free methods that directly optimize the policy over the dataset, model-based offline RL approaches begin by building a supervised transition model to provide pseudo-exploration around the offline logging dataset \cite{chua2018deep}. The agent can then interact with the transition model to optimize policy costlessly. Therefore, model-based methods are potentially more sample efficient than model-free methods. To avoid extrapolation error induced by the distributional shift, MOPO \cite{yu2020mopo} first proposed to penalize the reward with the uncertainty of transition model. They showed that policy performance could be guaranteed if the uncertainty is accurately assessed. However, the assessment is difficult since real-world state-action distributions are unavailable \cite{yu2021combo,panaganti2022robust}. Another line of work has tried to construct a pessimistic MDP to limit the occurrence of excessive uncertainty, such as MOReL \cite{kidambi2020morel}, CPPO \cite{uehara2021pessimistic} and ROMI \cite{wang2021offline}. DICE \cite{shen2021model} proposes to measure and penalty the extrapolation error by a density neural network. RAMBO \cite{rigter2022rambo} in a different way utilizes adversarial learning to restrict value estimation and force the policy conservatively. The difference is that our method explicitly introduces an adversary that can improve the generalization and penalize the reward at the same time. Despite the progress made by existing works, the transition models can easily overfit the offline logging dataset, and existing approaches tend to be over-conservative, making it challenging for the agent to efficiently utilize out-of-distribution regions for policy optimization. To address this, our method estimates uncertainty by introducing an adversary that learns the distance between the in-distribution and out-of-distribution samples. By doing so, the transition model can generate more diverse samples with accurate uncertainty estimation. This improves the agent's ability to utilize the out-of-distribution regions for policy optimization.

\section{Preliminaries}
\paragraph{Markov Decision Process} In RL, the environment is often modeled as a Markov Decision Process (MDP), defined by a quintuple $\mathcal{M} = (\mathcal{S},\mathcal{A}, R, \mathcal{T},\gamma)$. 
At each step $t$, when observing a state $s_t \in \mathcal{S}$, an agent will take action  $a_t\in \mathcal{A}$ according to the policy function $\pi(a_t|s_t)$. It will then receive an immediate reward $r(s_t,a_t) \in R$. The environment will transit to the next state $s_{t+1} \in \mathcal{S}$ according to the transition function $T(s_{t+1}, r_t|s_t, a_t) \in \mathcal{T}$. The agent aims to learn an optimal policy $\pi^*(\cdot)$ to maximize the $\gamma$-discounted cumulative reward: $\mathcal{J}(\pi, \mathcal{M}) = \mathbb{E}_{s\sim \mu_0}[\sum_{t=0}^{\infty}V_{\mathcal{M}}^{\pi}(s)]$, where $\mu_0$ is the initial state distribution and
$V_{\mathcal{M}}^{\pi}(s) = \mathbb{E}[\sum_{t=0}^{\infty}\gamma^tr(s_t,a_t)|s_0=s]$ is the expected value function from state $s$.

\begin{figure*}[tb]
    \centering
    \includegraphics[width=17.8cm]{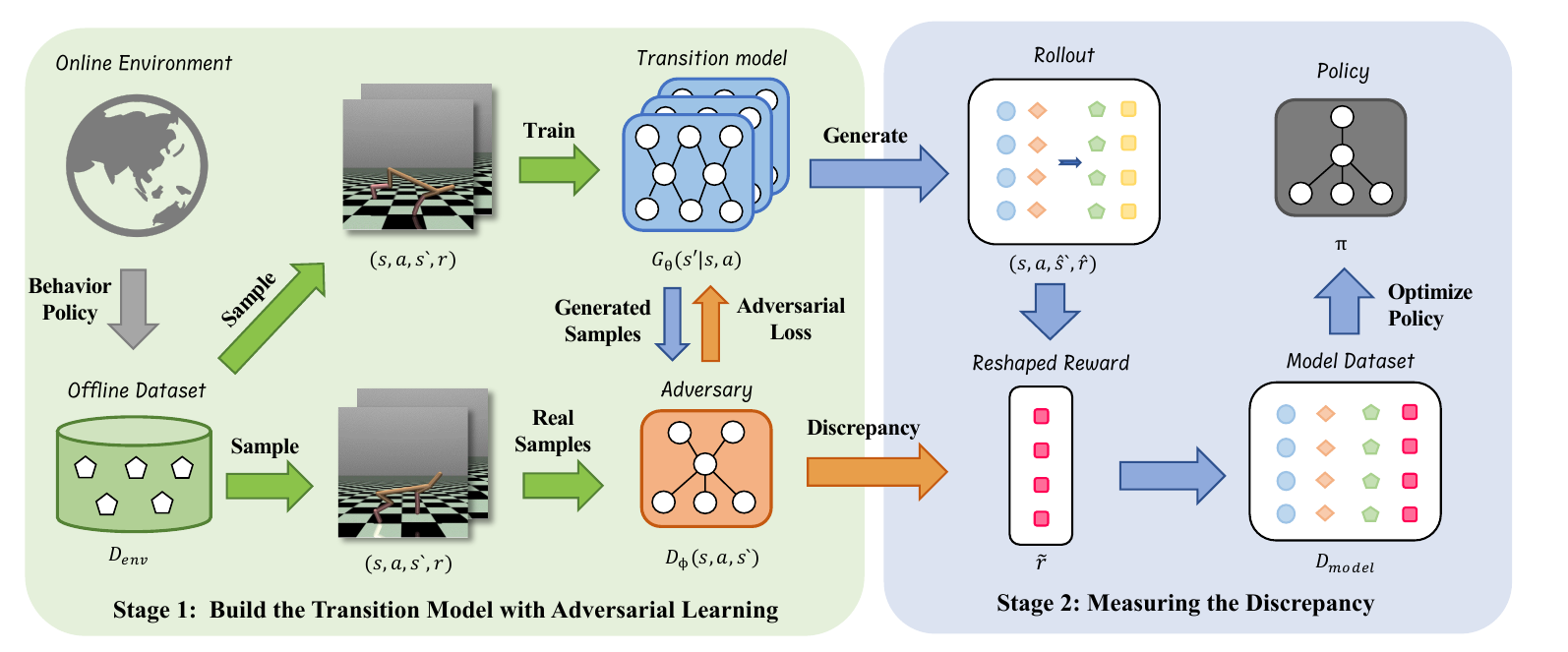}
    \caption{The framework of MOAN consists of two stages: In the first stage, the transition model is trained using adversarial learning to simulate the environment transitions. In the second stage, the agent samples rollouts from the transition model to optimize its policy. The distributional shift problem is avoided by correcting the prediction results using the adversary.}
    \label{fig:stru}
\end{figure*}

Let $\rho^\pi_{\mathcal{M}}(s,a) = (1-\gamma) \cdot \pi(a|s) \sum^\infty_{t=0}\gamma^t P^\pi_{\mathcal{M},t}(s)$ denote the normalized occupancy measure \cite{ho2016generative} for a policy $\pi$, where $P^\pi_{\mathcal{M},t}(s)$ represents the probability of state $s$ visited by $\pi$ under $\mathcal{M}$ at time step $t$. The discounted cumulative reward can be written as: $\mathcal{J}(\pi, \mathcal{M}) = \mathbb{E}_{(s, a)\sim\rho^{\pi}_{\mathcal{M}}}[r(s, a)]$. The normalized occupancy measure is utilized to compare the distribution of states and actions visited by different policies and dynamics.
\vspace{-3mm}
\paragraph{Model-based Offline RL} In offline RL, the agent cannot interact with the real environment, but only with a static logging dataset $\mathcal{D}_{env} = \{(s_i,a_i,s^\prime_i,r_i)\}^N_{i=1}$, collected by behavior policy $\pi_D$. To optimize the policy, model-based offline RL first builds a transition model $\hat{T}_\theta$ with supervised learning, which predicts the next state $s^{\prime}$ and reward $r$ based on the current state $s$ and action $a$. The transition model is often represented as a Gaussian distribution $\hat{T}_\theta(s^{\prime},r|s,a) = \mathcal{N}(\mu_\theta(s,a),\Sigma_\theta(s,a))$, where $\mu_\theta(s,a)$ is the mean of the distribution and $\Sigma_\theta(s,a)$ is the variance. The policy can then be optimized directly by interacting with the supervised transition model. Therefore, model-based methods are more sample efficient especially when the data collection is arduous. However, the difference between the real-world and the supervised transition model will accumulate extrapolation error and lead to the distribution shift problem. 

\section{Methodology}
In this section, we formally present our algorithm, named Model-based Offline Policy Optimization with Adversarial Network (MOAN). We begin by demonstrating how to build the transition model with adversarial learning. Next, we introduce a measurement of discrepancy between the real-world and the transition model to address the distributional shift in MOAN. The framework of our proposed method is illustrated in Figure~\ref{fig:stru}.

\subsection{Build the Transition Model with Adversarial Learning}
\label{Sec:4.1}
In the first stage of MOAN, we employ adversarial learning to train a transition model that generates diverse in-distribution samples. Specifically, we model the process as a two-player game, where one player is responsible for generating new samples, and the other adversary learns to distinguish the generated samples from the real ones. When converged, the first player is expected to generate samples that are indistinguishable from the offline dataset. In practice, we use a generative model $G$ to predict the next state, and a discriminator $D$ as the adversary to evaluate the authenticity of the generated samples.

More concretely, the transition model is implemented as an ensemble of $\mathcal{N}$ Gaussian distributions, denoted as $G_\theta = \{\hat{T}^1_\theta,..., \hat{T}^\mathcal{N}_\theta\}$, where $\theta$ is a parameter controlling the ensemble model. The transition model takes the current state-action pair $(s, a)$ as input and predicts the next state reward pair $(\hat{s}^{\prime}, \hat{r})$. The data sample ${(s, a,\hat{s}^{\prime}, \hat{r})}$ is then stored to the model's buffer $\mathcal{D}_{model}$, which is used to optimize offline policy. In this paper, we propose to maximize the confidence of the generated samples to improve the model's generalization. This is achieved by training an additional discriminator to estimate the confidence of the generated samples. Specifically, the confidence is defined as the probability that the discriminator assigns to the generated samples being real. By maximizing the confidence, we encourage the generator to produce samples that are not only indistinguishable from the real ones, but also have high confidence. This approach differs from maximum likelihood training \cite{yu2020mopo,kidambi2020morel}, which may easily overfit in the in-distribution dataset region. In contrast, our approach encourages the generator to explore more diverse regions of the state-action space, which can improve the model's generalization.

During the optimization process, the transition model $G_\theta$ and discriminator model $D_\phi$ are updated jointly. And the discriminator aims to distinguish between the generator predictions and the true labels. Formally, the discriminator is trained to minimize the probability of the generated samples while maximizing the probability of the real samples:
\begin{align}
    \label{eq:1}
    \mathcal{L}_{D}(\phi)=\mathbb{E}_{(s,a,s^\prime,r)\sim\rho_{\mathcal{M}}}[\log D_\phi(s,a,s^\prime,r)] \nonumber\\
    +\mathbb{E}_{(\hat{s},\hat{a},\hat{s}^\prime,\hat{r})\sim G_\theta}[\log (1 - D_\phi(\hat{s},\hat{a},\hat{s}^\prime,\hat{r}))].
\end{align}

For the generator, the loss function is a weighted sum of two parts: the first part is the negative log-likelihood loss of the transition model, and the second part is the adversarial loss:
\begin{align}
    \label{eq:2}
    \mathcal{L}_{G}(\theta)=\mathbb{E}_{(s,a,s^\prime,r)\sim\rho_{\mathcal{M}}}[-\log G_\theta(s^{\prime},r|s,a)] \nonumber\\
    +\alpha\mathbb{E}_{(\hat{s},\hat{a},\hat{s}^\prime,\hat{r})\sim G_\theta}[\log (1 - D_\phi(\hat{s},\hat{a},\hat{s}^\prime,\hat{r}))],
\end{align}
where $\alpha$ is a non-negative hyperparameter that balances the two parts. This hyperparameter can be tuned as a trade-off between the accuracy of the transition model and the diversity of the generated samples.

When the generator and discriminator have converged, the generator can produce data that closely resembles the in-distribution data. Moreover, this approach also maximizes the confidence from the discriminator, which encourages the generator to produce samples that are not only realistic but also diverse. By generating more diverse samples, the transition model can better explore the state-action space, especially in regions that are underrepresented in the in-distribution dataset. This leads to more effective policy optimization as the RL agent can utilize these diverse samples to learn better policies. Compared with previous model-based offline RL works, MOAN produces transition models with better generalization, which improves the agent's ability to exploit out-of-distribution regions of the state-action space.

\subsection{Measure the Discrepancy}
\label{Sec:4.2}
With the trained transition model, we can directly sample rollouts to optimize the policy of the RL agent. However, the discrepancy between the transition model and the real environment dynamics can lead to the failure of the policy. Measuring this discrepancy is difficult since the real environment dynamics can not be accessed. Nonetheless, the discriminator can output a probability value from $0 - 1$ to indicate whether the data is generated or collected from the real world to fix the distributional shift, and therefore provides a measure of the discrepancy. Formally, in the second stage of MOAN, we reshape the reward function as follows:
\begin{equation}
    \label{eq:r}
    \resizebox{0.90\linewidth}{!}{$
    \widetilde{r}(s,a) := r(s,a) - \eta(\Sigma_\theta(s,a) + \sqrt{ 2D_{\phi,(\hat{s}^\prime,\hat{r})\sim G(s,a)}(s,a,\hat{s}^\prime,\hat{r})})$},
\end{equation}
where $\eta$ is a hyperparameter to weight the reward penalty value. $\Sigma_\theta$ denotes the generator's variance and $D_\phi$ is the output of the discriminator. $\Sigma_\theta$ represents the uncertainty of the model's predictions, and can be regarded as the internal error of the transition model. $D_\phi$ represents the gap between the real-world model and the simulated transition model. The formulation indicates that we penalize the reward if the current state-action pairs have high uncertainty and low probability. Therefore, the policy of the agent can be constrained to the out-of-distribution region, to avoid the distributional shift problem.

The details of our algorithm are outlined in Algorithm \ref{algo:1}. As presented, we first jointly update a discriminator $D_\phi$ to distinguish the real and generated data, and a transition model $G_\theta$ with adversary learning (Line 1-6). Next, we generate new samples from $G_\theta$ (Line 12) and reshape the reward with the discrepancy described above (Line 13). The generated samples are then added to the model buffer $\mathcal{D}_{model}$ (Line 14). Finally, the policy can be optimized on the joined set of $\mathcal{D}_{env} \cup \mathcal{D}_{model}$ using a soft actor-critic (SAC) \cite{haarnoja2018soft} algorithm (Line 16). The overall goal of our method is to (1) update the transition model through adversarial learning to create a more accurate and comprehensive model and (2) optimize the policy using adversarial network penalty to correct the reward deviations, thereby reducing the distribution shift and improving the performance of offline RL.

\begin{algorithm}[!htb]
    \caption{Model-based Offline Policy Optimization with Adversarial Network (MOAN)}
    \renewcommand{\algorithmicrequire}{\textbf{Input:}}
	\renewcommand{\algorithmicensure}{\textbf{Output:}}
    \label{algo:1}
    \begin{algorithmic}[1]
        \REQUIRE Offline dataset $\mathcal{D}_{env}$, generator parameters $\theta$, discriminator parameters $\phi$, learning rate $\beta$ and $\omega$.
        \ENSURE Optimized policy  $\pi$.
        \STATE $\rhd$ building the transition model
        \WHILE{$\theta$ not converged} 
        \STATE Sample $(s,a,s^\prime,r)$ from $\mathcal{D}_{env}$.
        \STATE Update generator $\theta \leftarrow \theta + \beta \cdot \nabla_\theta \mathcal{L}_G$.
        \STATE Update discriminator $\phi \leftarrow \phi + \omega \cdot \nabla_\phi \mathcal{L}_D$.
        \ENDWHILE
        \STATE $\rhd$ model-based policy optimization
        \FOR{$m$ epochs}
        \STATE Sample an initial state $s_0$ from $\mathcal{D}_{env}$. 
        \FOR{$h$ horizon}
        \STATE Sample an action $a \sim \pi(s)$.
        \STATE Sample $s^\prime,r \sim G_\theta(s,a)$.
        \STATE Reshape the reward $\widetilde{r}$ according to Eq.~(\ref{eq:r}).
        \STATE Add samples $(s,a,s^\prime,\widetilde{r})$ to model buffer $\mathcal{D}_{model}$.
        \ENDFOR 
        \STATE Optimize $\pi$ with SAC from $\mathcal{D}_{env} \cup \mathcal{D}_{model}$.
        \ENDFOR
    \end{algorithmic}
\end{algorithm}

\section{Theoretical Analysis}
\label{sec:4.3}
In this section, we analyze the performance of MOAN theoretically. In particular, we show that the expected return of the policy $\pi$ in the real environment $\mathcal{J}(\pi, \mathcal{M})$ is at least as high as the expected return of the policy $\pi$ in the simulated environment $\mathcal{J}(\pi, \hat{\mathcal{M}})$ plus a bias item, where $\hat{\mathcal{M}}$ represents the simulated MDP in Section 4.1. The lower bound can be provided with the following theorem:

\begin{theorem}
\label{thorem:1}
Let $T(\cdot|s,a)$ and $\hat{T}(\cdot|s,a)$ be the transition functions of $\mathcal{M}$ and $\hat{\mathcal{M}}$ with the same reward \cite{yu2020mopo,luo2018algorithmic} and bounded value function. The lower bound of real environment expected return can be denoted as:
    \begin{align}
    \hspace{-2mm}
    &\mathcal{J}(\pi, \mathcal{M}) \geq \mathcal{J}(\pi, \hat{\mathcal{M}})\nonumber\\     
    \begin{split}
    &\resizebox{0.87\linewidth}{!}{$-\gamma(\underbrace{\mathbb{E}_{(s,a)\sim \rho^{\pi_D}_{\mathcal{M}}}[d_{TV}(T(\cdot|s,a),\hat{T}(\cdot|s,a))]}_{\text{transition model error}}+\underbrace{\sqrt{2d_{JS}(\rho^{\pi_D}_{\mathcal{M}},\rho^{\pi}_{\hat{\mathcal{M}}})})}_{\text{distribution discrepancy}}$},
    \end{split}
    \label{eq:4}
    \end{align}
    where $d_{TV}(\cdot)$ is the Total Variation (TV) distance \cite{devroye2018total} and $d_{JS}(\cdot)$ is the Jenson-Shanon (JS) divergence \cite{fuglede2004jensen}.
\end{theorem}
\textbf{\textit{proof.}} According to the definition of $\gamma$-discounted return mentioned in preliminaries, the difference between the real environment $\mathcal{M}$ and simulated environment $\mathcal{\hat{M}}$ returns can be formulated as:
\begin{align}
    \label{eq:theo1}
    &\mathcal{J}(\pi, \mathcal{M}) - \mathcal{J}(\pi,\hat{\mathcal{M}}) = \nonumber \\
    \begin{split}
    &\resizebox{0.87\linewidth}{!}{$\gamma \mathbb{E}_{(s,a)\sim \rho^{\pi}_{\hat{\mathcal{M}}}}[\mathbb{E}_{s^{\prime}\sim T(\cdot \mid s,a)}[V^{\pi}_\mathcal{M}(s^{\prime})]-\mathbb{E}_{s^{\prime}\sim \hat{T}(\cdot \mid s,a)}[V^{\pi}_\mathcal{M}(s^{\prime})]]$}.
    \end{split}
\end{align}
To simplify the notations, we can use the value estimation discrepancy between real transition $T(\cdot | s,a)$ and simulated transition $\hat{T}(\cdot | s,a)$ to substitute Eq.\ref{eq:theo1}. $Z^{\pi}_{\hat{\mathcal{M}}}(s,a):=\mathbb{E}_{s^{\prime}\sim \hat{T}(\cdot | s,a)}[V^{\pi}_\mathcal{M}(s^{\prime})] - \mathbb{E}_{s^{\prime}\sim T(\cdot | s,a)}[V^{\pi}_\mathcal{M}(s^{\prime})]$. Then We have a more concise expression:
\begin{equation}
    \mathcal{J}(\pi, \mathcal{M}) - \mathcal{J}(\pi,\hat{\mathcal{M}})=-\gamma \mathbb{E}_{(s,a)\sim \rho^{\pi}_{\hat{\mathcal{M}}}}[Z^{\pi}_{\hat{\mathcal{M}}}(s,a)].
\end{equation}
Since we aim to measure the state-action distribution deviation between two environments and the normalized occupancy distribution $\rho_{\mathcal{M}}^\pi$ from the real environment by a training policy is not directly accessible, we introduce $\rho_{\mathcal{M}}^{\pi_D}$ from a offline dataset $\mathcal{D}_{env}$, as Eq.\ref{eq:7} denoted.
\begin{equation}
\begin{aligned}
&\mathcal{J}(\pi, \mathcal{M}) - \mathcal{J}(\pi,\hat{\mathcal{M}})\\
=&-\gamma \mathbb{E}_{(s,a)\sim \rho^{\pi}_{\hat{\mathcal{M}}}}[Z^{\pi}_{\hat{\mathcal{M}}}(s,a)] + \gamma \mathbb{E}_{(s,a)\sim \rho^{\pi_D}_{\mathcal{M}}}[Z^{\pi}_{\hat{\mathcal{M}}}(s,a)]\\
\,&- \gamma \mathbb{E}_{(s,a)\sim \rho^{\pi_D}_{\mathcal{M}}}[Z^{\pi}_{\hat{\mathcal{M}}}(s,a)].
\label{eq:7}
\end{aligned}
\end{equation}
Furthermore, the value estimation discrepancy $Z^{\pi}_{\hat{\mathcal{M}}}(s,a)$ can be constrained by integral probability metric (IPM)\cite{muller1997integral} that the value functions are limited within the function collection $\mathcal{F} =\{ f: \mathcal{S} \times \mathcal{A} \to \mathbb{R} \big|  \Vert f \Vert_{\infty} \leq \delta \}$. And we have: 
\begin{equation}
\begin{aligned}
    Z^{\pi}_{\hat{\mathcal{M}}}(s,a) &\leq \sup_{f\in \mathcal{F}}\left|\mathbb{E}_{s^{\prime}\sim T(\cdot \mid s,a)}[f(s^\prime)]-\mathbb{E}_{s^{\prime}\sim \hat{T}(\cdot \mid s,a)}[f(s^\prime)] \right|\\
    &=:d_{f\in \mathcal{F}}(\hat{T}(\cdot | s,a), T(\cdot | s,a)),
\end{aligned}
\end{equation}
\begin{table*}[htbp]
    \begin{center}
    \begin{tabular}{ccccccccccc}
       \toprule
       \textbf{Dataset type} & \textbf{Environment} & \textbf{MOAN}  & \textbf{MOPO} & \textbf{MOReL} & \textbf{RAMBO} & \textbf{COMBO} & \textbf{CQL}   & \textbf{IQL}  & \textbf{TD3+BC} & \textbf{BC} \\
       \midrule
random                & halfcheetah          & $\bm{39.2 \pm 5.2}$  & 35.4 & 25.6           & \textbf{39.5}  & \textbf{38.8}  & 19.6           & -             & 11.0              & 2.1         \\
random                & hopper               & $31.2 \pm 3.1$           & 11.7          & \textbf{53.6}  & 25.4           & 17.9           & 6.7            & -             & 8.5             & 9.8         \\
random                & walker2d             & $18.2 \pm 5.6$          & 13.6          & \textbf{37.3}  & 0.0              & 7.0              & 2.4            & -             & 1.6             & 1.6         \\
medium                & halfcheetah          & $63.5 \pm 4.7$          & 42.3          & 42.1           & \textbf{77.9}  & 54.2           & 49.0             & 47.4          & 48.3            & 36.1        \\
medium                & hopper               & $\bm{101.3 \pm 4.7}$ & 28.0            & 95.4  & 87.0             & \textbf{97.2}  & 66.6           & 66.3          & 59.3            & 29.0          \\
medium                & walker2d             & $\bm{89.7 \pm 1.1}$  & 17.8          & 81.9  & 84.9  & 54.2           & 83.8  & 78.3          & 83.7   & 6.6         \\
med-replay            & halfcheetah          & $\bm{65.3 \pm 4.4}$  & 53.1          & 40.2           & \textbf{68.7}  & 55.1           & 47.1           & 44.2          & 44.6            & 38.4        \\
med-replay            & hopper               & $\bm{103.9 \pm 3.3}$ & 67.5          & 93.6           & \textbf{99.5}  & 89.5           & 97    & 94.7 & 60.9            & 11.8        \\
med-replay            & walker2d             & $78.1 \pm 6.5$  & 39.0            & 49.8           & \textbf{89.2}  & 56.0             & \textbf{88.2}  & 73.9          & 81.8   & 11.3        \\
med-expert            & halfcheetah          & $\bm{95.4 \pm 2.3}$  & 63.3          & 53.3           & \textbf{95.4}  & \textbf{90.0}    & \textbf{90.8}  & 86.7          & \textbf{90.7}   & 35.8        \\
med-expert            & hopper               & $\bm{108.3 \pm 8.7}$ & 23.7          & \textbf{108.7} & 88.2           & \textbf{111.1} & \textbf{106.8} & 91.5          & 98.0              & \textbf{111.9}       \\
med-expert            & walker2d             & $83.4 \pm 12.7$           & 44.6          & 95.6           & 56.7           & 103.3 & \textbf{109.4} & 109.6         & \textbf{110.1}  & 6.4        \\
       \bottomrule
    \end{tabular}
    \end{center}
    \caption{Performance of MOAN and prior baselines on the D4RL tasks. Scores are the average returns value with 10 evaluations in the real environment, averaged over 5 random seeds. All values are normalized to lie between 0 and 100, where 0 corresponds to a random policy and 100 corresponds to an expert policy \protect\cite{fu2020d4rl}. And baseline values were taken from their respective papers. Boldface denotes performance within 5\% of the best performing algorithm.}
    \label{tab:d4rl}
\end{table*}
where $d_{f\in \mathcal{F}}(\cdot)$ is the IPM defined by the function class $\mathcal{F}$. By choosing different categories of measurement $\mathcal{F}$, IPM can reduce to many different well-known distance metrics between probability distributions, such as Total Variation (TV) distance \cite{devroye2018total} and maximum mean discrepancy (MMD) \cite{gretton2012kernel}. 
\begin{equation}
\begin{aligned}
&\mathcal{J}(\pi, \mathcal{M}) - \mathcal{J}(\pi,\hat{\mathcal{M}})\\
\geq&- (\gamma \mathbb{E}_{(s,a)\sim \rho^{\pi}_{\hat{\mathcal{M}}}}[Z^{\pi}_{\hat{\mathcal{M}}}(s,a)] - \mathbb{E}_{(s,a)\sim \rho^{\pi_D}_{\mathcal{M}}}[Z^{\pi}_{\hat{\mathcal{M}}}(s,a)])\\
~\,&-\gamma \mathbb{E}_{(s,a)\sim \rho^{\pi_D}_{\mathcal{M}}}[d_{f\in \mathcal{F}}(\hat{T}(\cdot | s,a), T(\cdot | s,a))]\\
=&-\gamma d_{TV}(\rho^{\pi_D}_{\mathcal{M}}, \rho^{\pi}_{\hat{\mathcal{M}}})\\
~\,&-\gamma \mathbb{E}_{(s,a)\sim \rho^{\pi_D}_{\mathcal{M}}}[d_{TV}(\hat{T}(\cdot | s,a), T(\cdot | s,a))],
\end{aligned}
\end{equation}
where the distance $d_{f\in \mathcal{F}}(\cdot)$ is implemented as the TV distance in the second equation.

Since it is difficult to learn the state-action distribution $\rho^\pi_{\hat{\mathcal{M}}}$, we use Pinsker's Inequality \cite{fedotov2003refinements} to transform the squared TV distance into a KL divergence equation. According to the Jenson-Shanon (JS) divergence definition, we can convert the the sum of the KL divergence into a GAN-like objective that the can be directly learned by a neural network:
\begin{equation}
    \begin{aligned}
    &\quad d_{TV}(\rho^{\pi_D}_{\mathcal{M}}, \rho^{\pi}_{\hat{\mathcal{M}}}) \\
    =&\sqrt{2(d_{TV}^2(\rho^{\pi_D}_{\mathcal{M}},\frac{\rho^{\pi_D}_{\mathcal{M}}+\rho^{\pi}_{\hat{\mathcal{M}}}}{2})+d_{TV}^2(\rho^{\pi}_{\hat{\mathcal{M}}},\frac{\rho^{\pi_D}_{\mathcal{M}}+\rho^{\pi}_{\hat{\mathcal{M}}}}{2}))}\\
    \leq &\sqrt{(d_{KL}(\rho^{\pi_D}_{\mathcal{M}},\frac{\rho^{\pi_D}_{\mathcal{M}}+\rho^{\pi}_{\hat{\mathcal{M}}}}{2})+d_{KL}(\rho^{\pi}_{\hat{\mathcal{M}}},\frac{\rho^{\pi_D}_{\mathcal{M}}+\rho^{\pi}_{\hat{\mathcal{M}}}}{2}))}\\
    =&\sqrt{2d_{JS}(\rho^{\pi_D}_{\mathcal{M}},\rho^{\pi}_{\hat{\mathcal{M}}})}.
    \end{aligned}
\end{equation}

In Theorem~\ref{thorem:1}, the lower bound of the expected return in the real environment can be represented as a combination of 3 terms: the expected returns of $\pi$ in $\hat{\mathcal{M}}$, the transition model error, and the distribution discrepancy. The second term represents the error between the real-world transition model $T$ and the simulated transition model $\hat{T}$. The third term measures the discrepancy between the model rollout data distribution $\rho^{\pi}_{\hat{\mathcal{M}}}$ and the offline data distribution $\rho^{\pi_D}_{\mathcal{M}}$. Compared with prior methods \cite{yu2020mopo,kidambi2020morel}, we additionally introduce this discrepancy term to ensure that the model rollout distribution is closer to the true distribution. Our bound is tighter than MOPO. As presented in Eq. \ref{eq:4}, MOPO did not consider the distribution discrepancy term. Considering this term can potentially reduce the risks of exploring out-of-distribution data.

Essentially, the theorem indicates that if the transition model is accurate and the state-action distributions are similar (as shown in Sec~\ref{Sec:4.1}), and the distribution discrepancy can be precisely measured by the adversary (as shown in Sec~\ref{Sec:4.2}), then the resulting performance of the policy in the real environment will be guaranteed.

\section{Experiments}
In this section, we evaluate the effectiveness of the proposed MOAN to answer the following four questions\footnote{Code and appendix is available at \href{https://github.com/junming-yang/MOAN}{https://github.com/junming-yang/MOAN}}: (1) How is the performance of MOAN compared to state-of-the-art offline RL algorithms on typical benchmark tasks? (2) Can MOAN generate diverse samples that closely mimic the online environment? (3) Can MOAN accurately measure the discrepancy between the real world and the simulated transition model? (4) What is the performance of offline policy optimization with the different hyperparameters for adversarial generation and discrepancy measuring?

\subsection{Experiment Setting}
Our experiments are conducted in three MuJoCo simulated environments (HalfCheetah, Hopper and Walker2d) using a standard offline RL benchmark D4RL \cite{fu2020d4rl}. Each environment has four offline logging datasets, collected by different one or mixed behavior policies: \textit{random, medium, medium-replay, and medium-expert}. The \textit{random} dataset is created by implementing a randomly generated policy for $10^6$ steps. The \textit{medium} dataset is generated by using a soft actor-critic policy that has been trained to achieve approximately 1/3 of the performance level of an expert. The \textit{medium-replay} dataset is made up of all the samples in the replay buffer during the training process until the policy reaches the medium level of performance. The \textit{medium-expert} dataset is created by combining an equal number of expert-level samples and medium-level samples. All results are evaluated 1000 episode (1M steps) runs per seed in real environment, and averaged based on five random seeds.
More details on the experimental setup, different environmental hyperparameter configurations, generator and discriminator network structures, and codes are attached to Appendix A.

To thoroughly evaluate the performance of our approach on the D4RL dataset, we conduct a series of experiments and compare the results to several offline RL baselines. These baselines include Behavior Cloning method (BC) and model-free methods Constrain Q-learning (CQL) \cite{kumar2020conservative}, Implicit Q-learning (IQL) \cite{kostrikov2021offline}, and Minimalist Offline RL Algorithm (TD3+BC) \cite{fujimoto2021minimalist}, as well as Model-based offline policy optimization (MOPO) \cite{yu2020mopo}, model-based methods Model-based offline reinforcement learning (MOReL) \cite{kidambi2020morel}, Robust Adversarial Model-Based Offline Reinforcement Learning(RAMBO) \cite{rigter2022rambo} and Conservative offline 
model-based policy optimization (COMBO) \cite{yu2021combo}. 

\begin{figure*}[tb]
\begin{center}
\subfigure[Hopper Medium-replay Env.]{
\includegraphics[width=5.5cm]{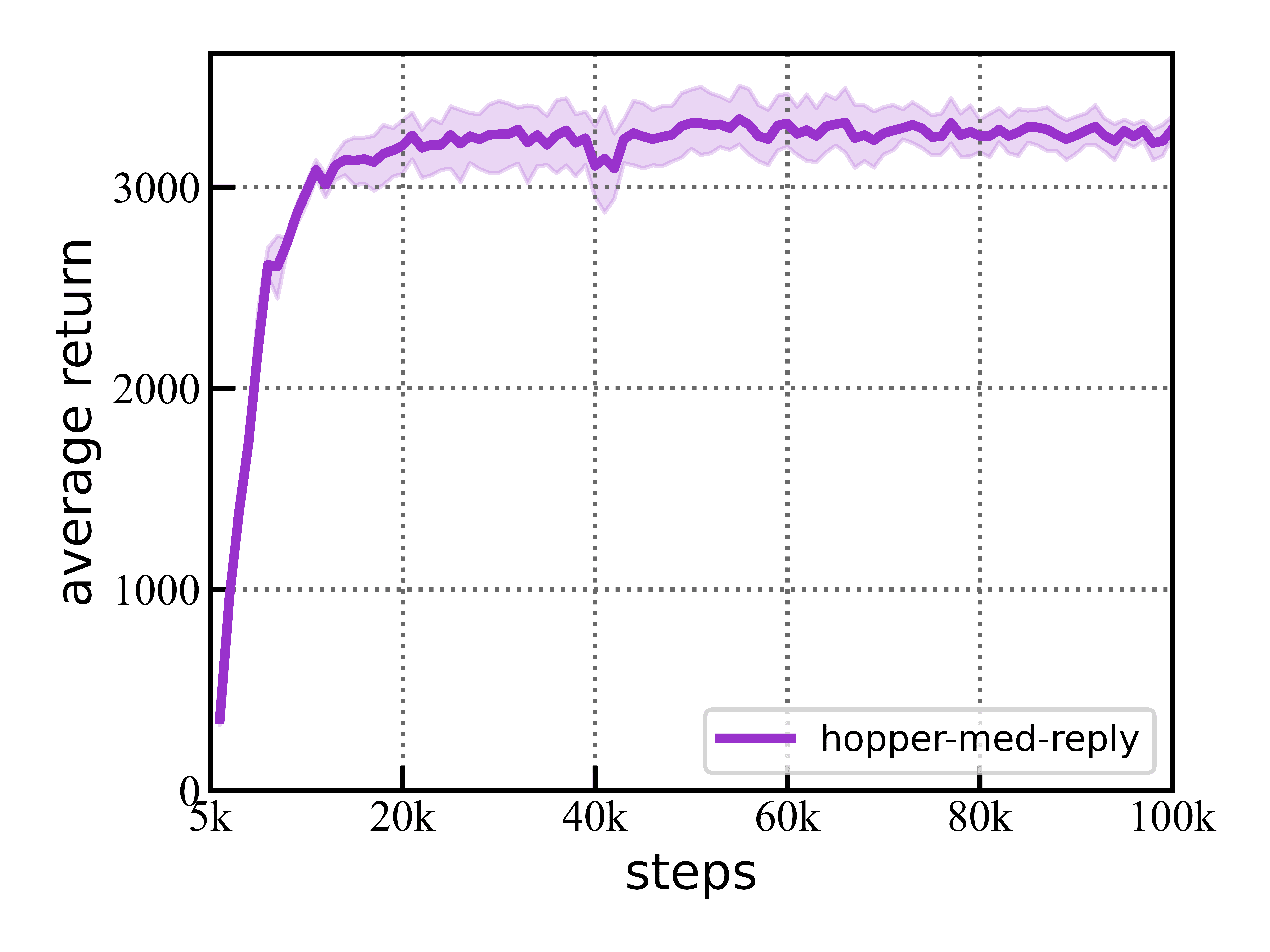}
}
\subfigure[Halfcheetach Medium-replay Env.]{
\includegraphics[width=5.5cm]{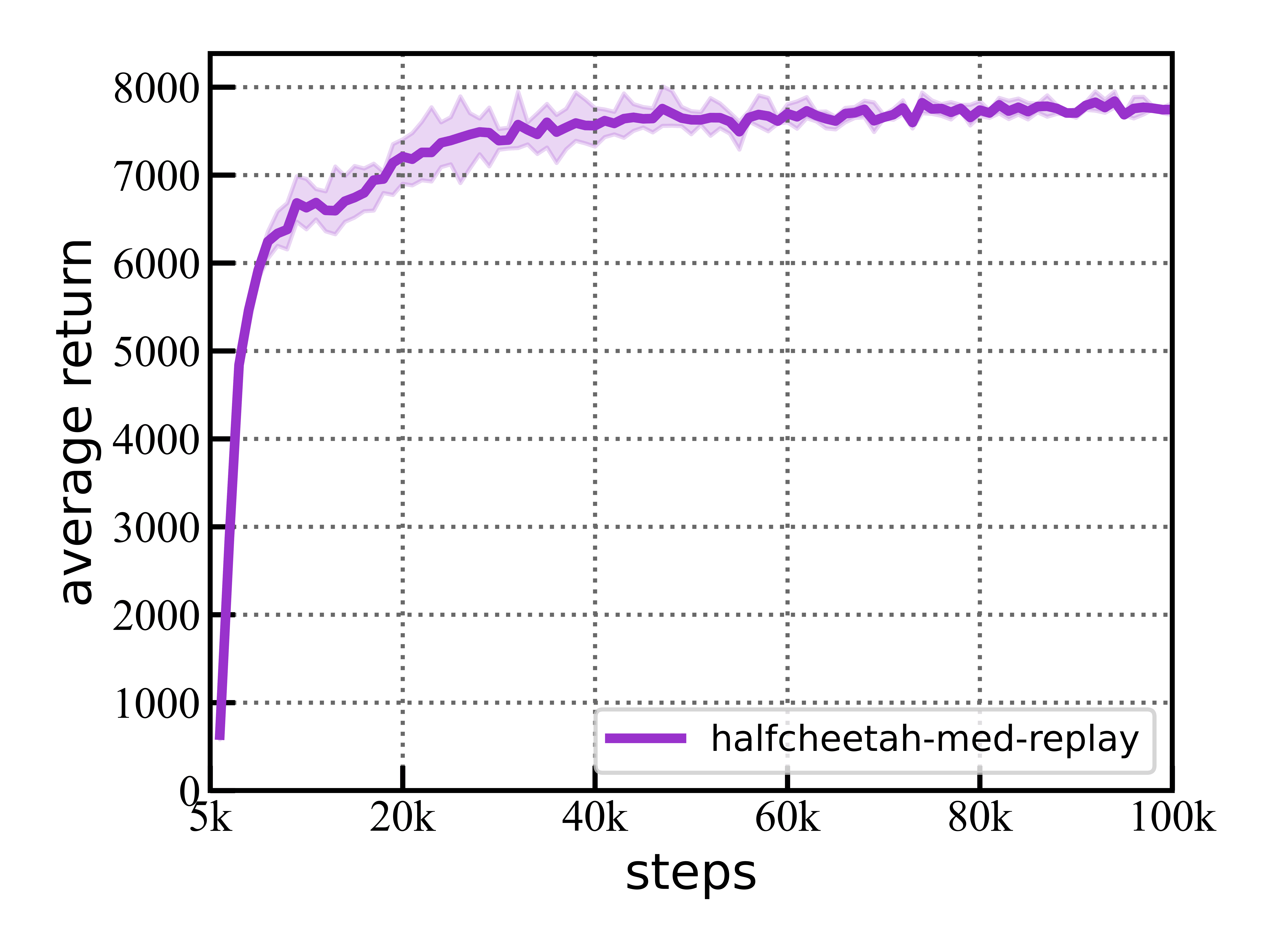}
}
\subfigure[Walker2d Medium-replay Env.]{
\includegraphics[width=5.5cm]{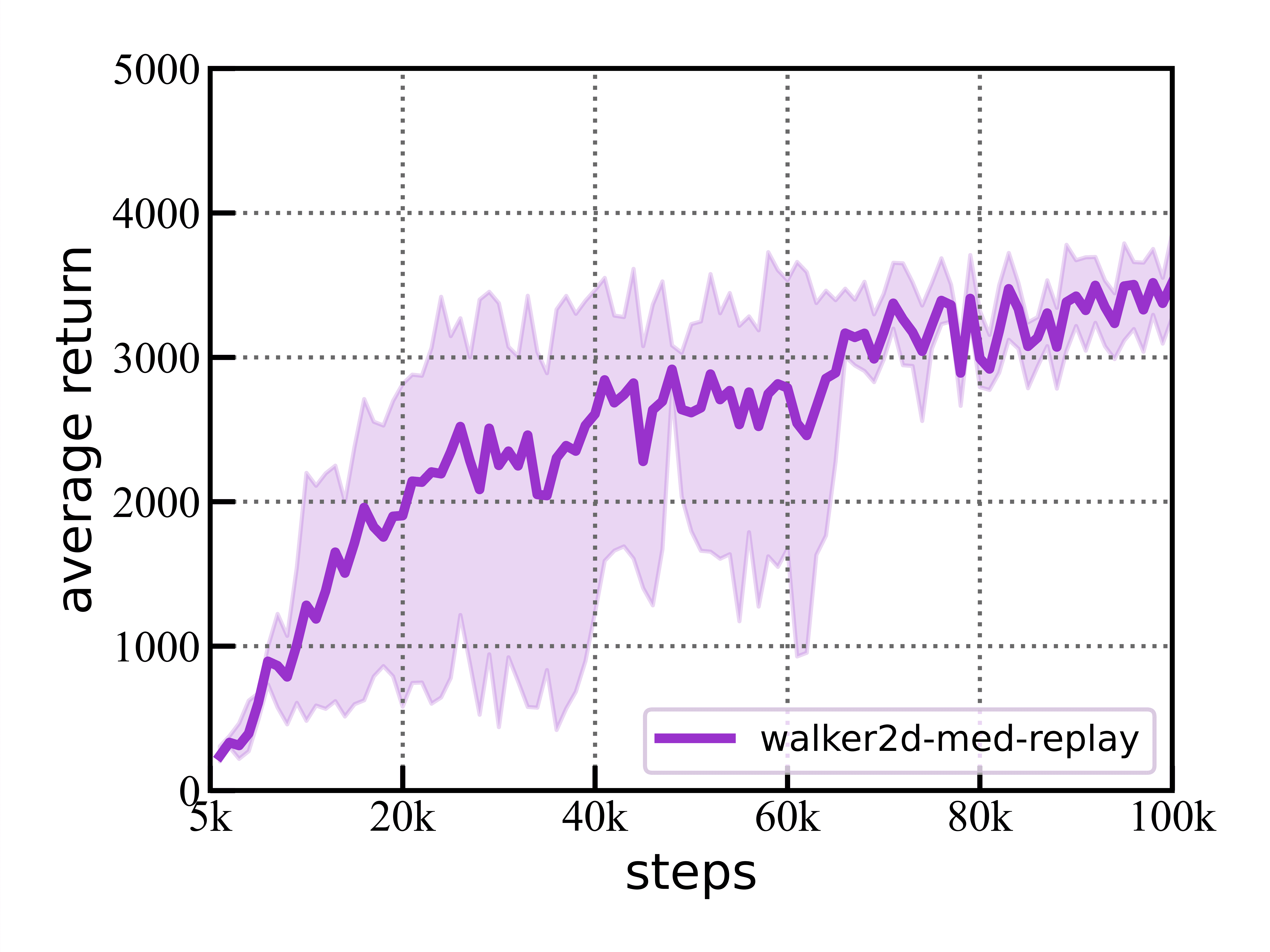}
}
\caption{D4RL Benchmark Result. (a) The learning curves of the three algorithms MOAN on the Hopper environment \textit{medium-replay} setting. (b) The learning curves of the three algorithms MOAN on the Halfcheetah environment \textit{medium-replay} setting. (c) The learning curves of the three algorithms MOAN on the Walker2d environment \textit{medium-replay} setting. The curve represents the mean score and the shaded part represents the statistical standard deviation under multiple random seed experiments.}
\label{fig:ab}
\end{center}
\vspace{-3mm}
\end{figure*}

\subsection{Evaluation on D4RL Benchmark}
First, we evaluate the expected returns of different methods on a set of D4RL benchmarks with several settings. The evaluation results are presented in Table~\ref{tab:d4rl}. As presented, MOAN achieves the best performance on 7 tasks across all 12 dataset settings, and achieves comparable results (the second best) in other 3 out of 5 datasets. In particular, MOAN has superior performance in the Hopper environment. This is because Hopper has lower dimensions of states and actions, which may lead to over-conservative in other algorithms.

Compared with the model-based algorithms, MOAN can outperform MOPO across all settings. MOReL has better performance in two \textit{random} settings. The reason is that MOReL adopts a more conservative policy, which reduces the occurrence of dangerous situations in the random datasets. RAMBO performs well on \textit{med-replay} datasets because its transition model also has a certain generalization. However, due to its inability to accurately correct the extrapolation error, RAMBO adopted a conservative approach and had a mediocre effect on other datasets. Compared with COMBO, which is the state-of-the-art model-based method, MOAN still has higher performance in 9 out of the 12 settings. When compared to model-free methods such as CQL, IQL, and TD3+BC, MOAN has better performance in almost all settings, particularly in the \textit{random} and \textit{medium} datasets. This is because MOAN can utilize offline datasets more efficiently by building the transition model. Overall, our evaluation results provide a clear indication of the strengths and effectiveness of the MOAN method in offline RL.

We also compare the average return of MOAN, MOPO and CQL in the Hopper environment during the training phase. MOPO has a similar model structure to MOAN, and CQL is the state-of-the-art algorithm in this setting. The learning progresses are presented in Figure \ref{fig:ab}. The curves are the mean value of the average returns from multiple runs, and the shaded areas represent the standard derivations. Obviously, MOAN can achieve the best performance across the training phase. Moreover, the results show that MOAN can converge very fast in only about 20k steps.

\subsection{Performance of the Transition Model}
To evaluate the effectiveness of the transition model in our MOAN algorithm, we compared its performance with that of MOPO, which has a similar transition model structure. For each task, we trained the transition models independently until convergence, and then randomly generated 10,000 samples from the real environment. These samples were fed into the different transition models to predict the next states.

\begin{figure}[tb]
\begin{center}
\subfigure[MOPO algorithm.]{
\includegraphics[width=4.05cm]{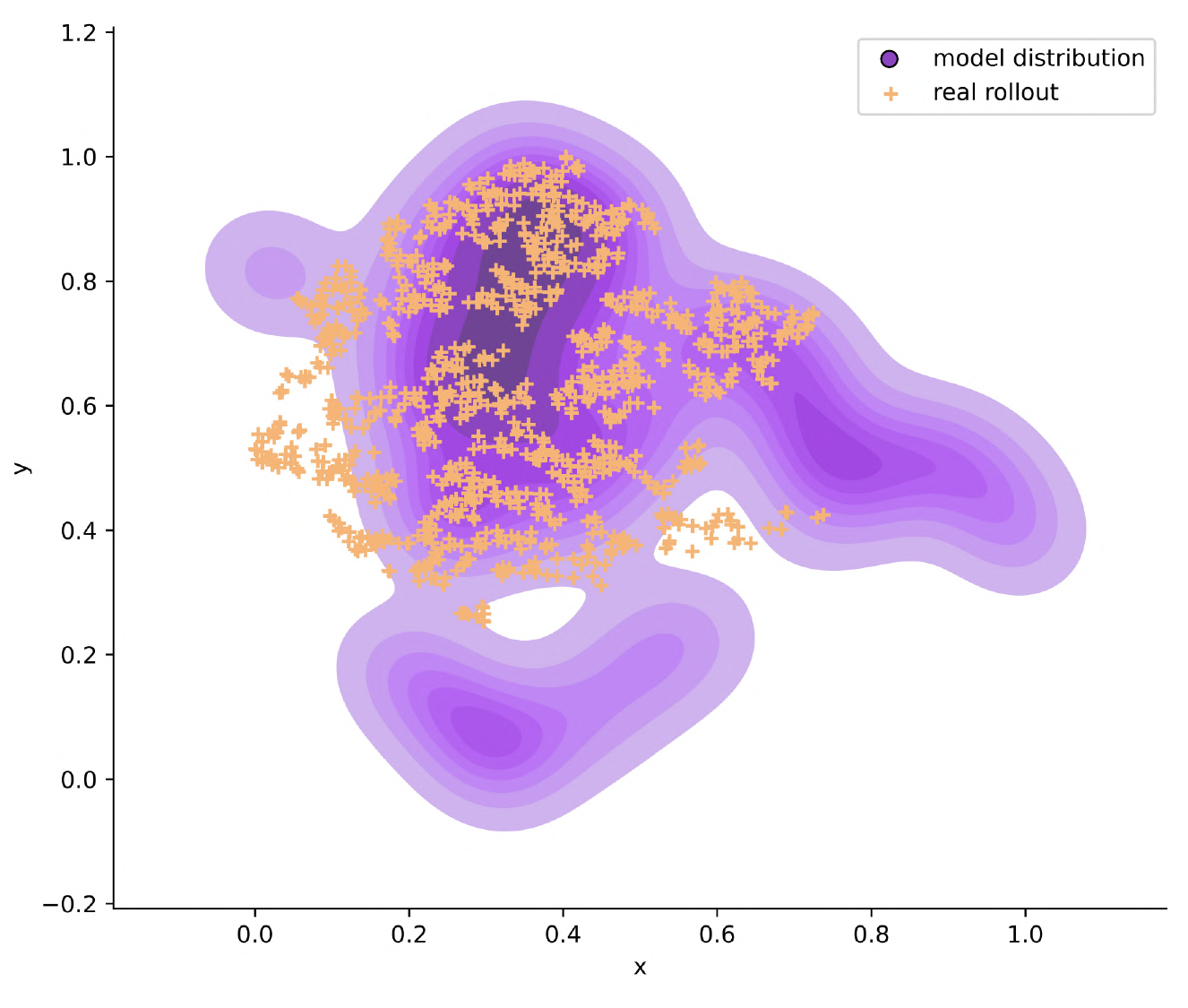}
}
\subfigure[MOAN algorithm.]{\includegraphics[width=4.05cm]{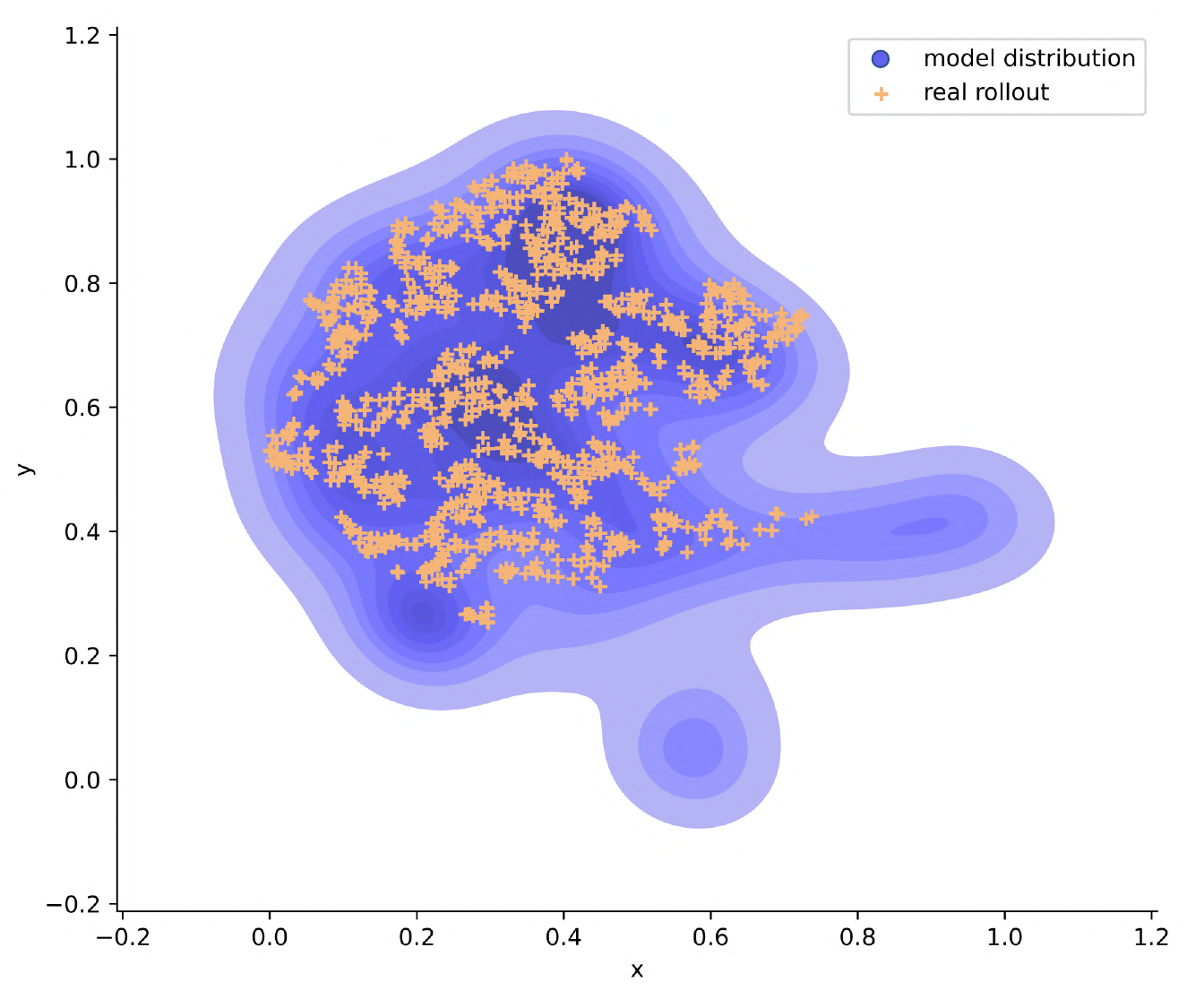}
}
\caption{Visualization of the two-dimensional t-SNE state distribution learned by transition model from hopper environment. Colors correspond to different datasets. Purple: MOPO model state distribution, Blue: MOAN model state distribution, Yellow crosses: real-world rollout samples. The dark colors represent areas where model samples distribution is concentrated.}
\label{fig:tsne}
\end{center}
\end{figure}

We visualized the state distributions using t-SNE and presented the results in Figure~\ref{fig:tsne}. The yellow crosses represent the rollouts from the real environment, while the model states generated by MOPO and MOAN are colored purple and red, respectively. To facilitate comparison, we drew the results of model prediction as distribution maps, where the darker parts represent more concentrated regions of the results.

Our results show that the real rollout points in MOPO are more likely to be outside of the model distribution, while they are always consistent with the model distribution in MOAN. This indicates that MOAN not only generates more similar states that are within the real dataset but also has a better generalization to cover the distribution. These findings suggest that the transition model of MOAN has better generalization and accuracy compared to that of the MOPO algorithm.

\begin{figure}[tb]
\begin{center}
\subfigure[MOPO algorithm.]{
\includegraphics[width=4.05cm]{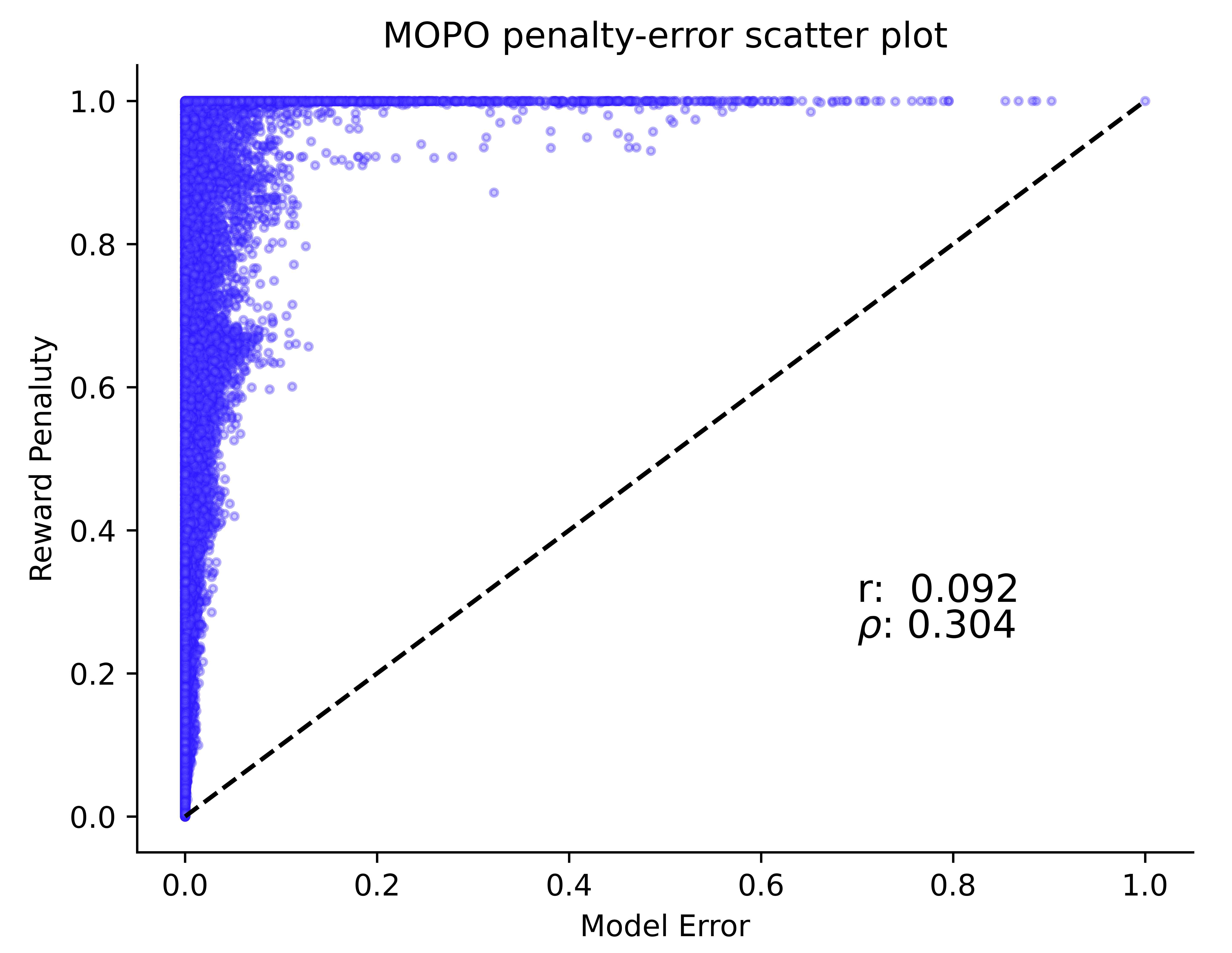}
}
\subfigure[MOAN algorithm.]{
\includegraphics[width=4.05cm]{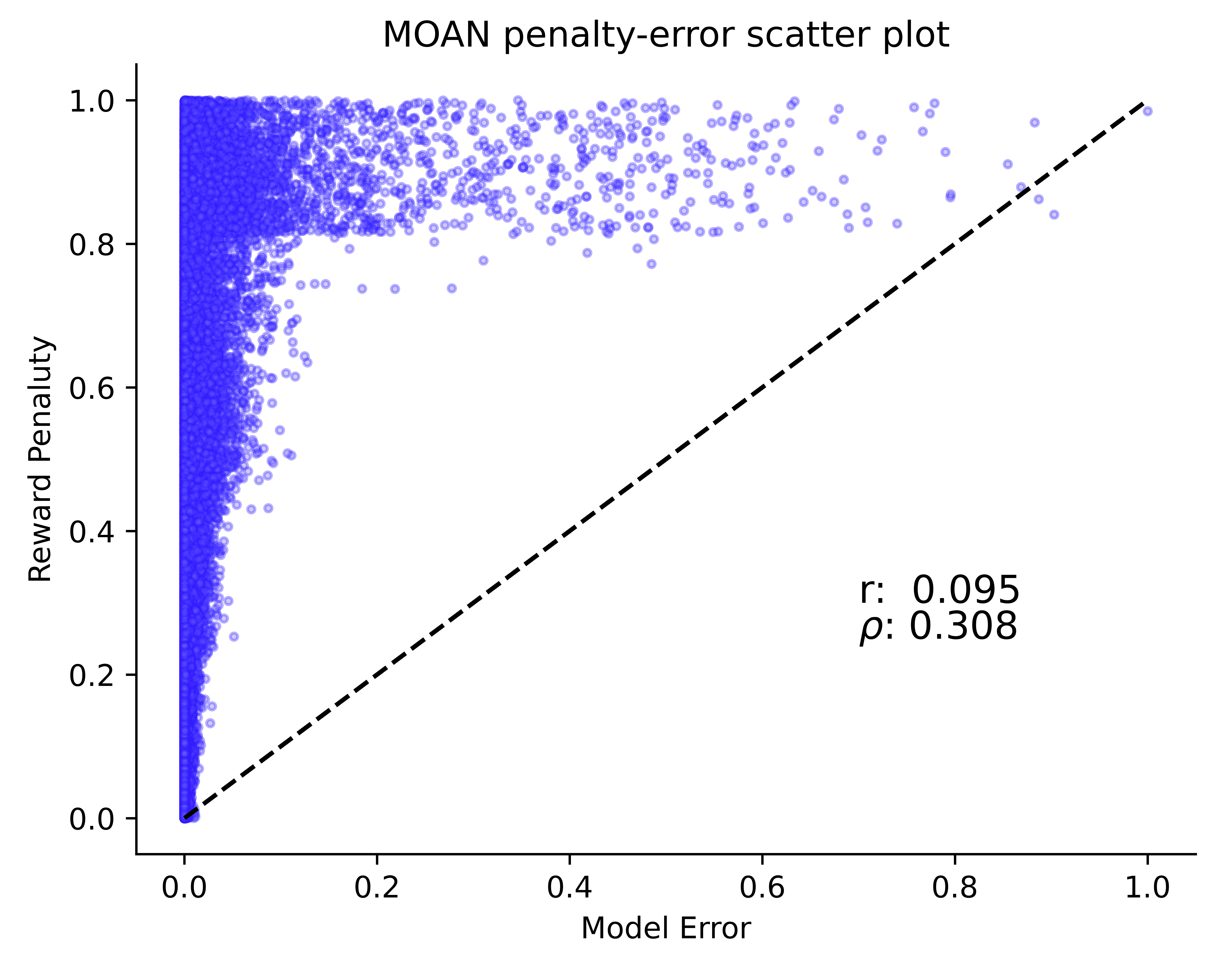}
}
\caption{Visualization of the correlation between the ground-truth MSE and reward penalty values. The reward penalty values are predicted from action-state pairs that are randomly selected from the dataset. The left figure shows the results of the MOPO method, while the right figure shows the results of the MOAN method.}
\label{fig:linear}
\end{center}
\end{figure}

\subsection{Performance of the Discrepancy Measure} 
We conducted experiments to investigate whether our proposed MOAN algorithm can more accurately quantify the discrepancy between an offline dataset and the real environment. Specifically, we compared the results in the Hopper environment using the \textit{medium-replay} dataset. The primary objective of this experiment was to learn more about how accurately both algorithms could quantify discrepancies in the real-world environment, particularly the accuracy of the reward penalties given by each algorithm. We visualized the results in Figure~\ref{fig:linear}, where the x-axis represents the mean squared error (MSE) of the ensemble model $\Vert \hat{T}(\cdot|s,a) - T(\cdot|s,a) \Vert_2$, and the y-axis represents the normalized reward penalty value. We normalized the reward penalty and model errors on a per-dimension basis to the [0, 1] interval so that the scattered points should lie along the diagonal line $y = x$ in an ideal situation where the discriminator perfectly captures the simulated dynamics error.

As shown in Figure~\ref{fig:linear}, the results of the experiment prove effectiveness of MOAN, with the scattered points of MOAN located much closer to the diagonal line. In contrast, MOPO's scattered points were mainly located higher on the diagram, indicating an emphasis on over-conservatism and a higher reward penalty value. This result demonstrates that MOAN has greater accuracy in quantifying discrepancies in the real environment between the offline dataset.

Both MOAN and MOPO use penalty functions to encourage the online policy to explore the environment. However, MOPO uses a handcrafted penalty function, whereas MOAN designs a neural network-based penalty function that can learn to balance and optimize exploration and exploitation dynamically. MOAN uses the discriminator as the reward penalty function, which is also used in training the generator to eliminate the shift between the offline dataset and the real environment. 

\subsection{Ablation Study}
In the last part of our study, two ablation studies were conducted to investigate the contributions of adversarial learning and discrepancy penalty in MOAN in an online environment. The performance of the algorithm was evaluated with different values of adversarial learning hyperparameter $\alpha$ and discrepancy measurement hyperparameter $\eta$.

\begin{figure}[tb]
\begin{center}
\subfigure[Ablation on $\alpha$.]{
\includegraphics[width=4.05cm]{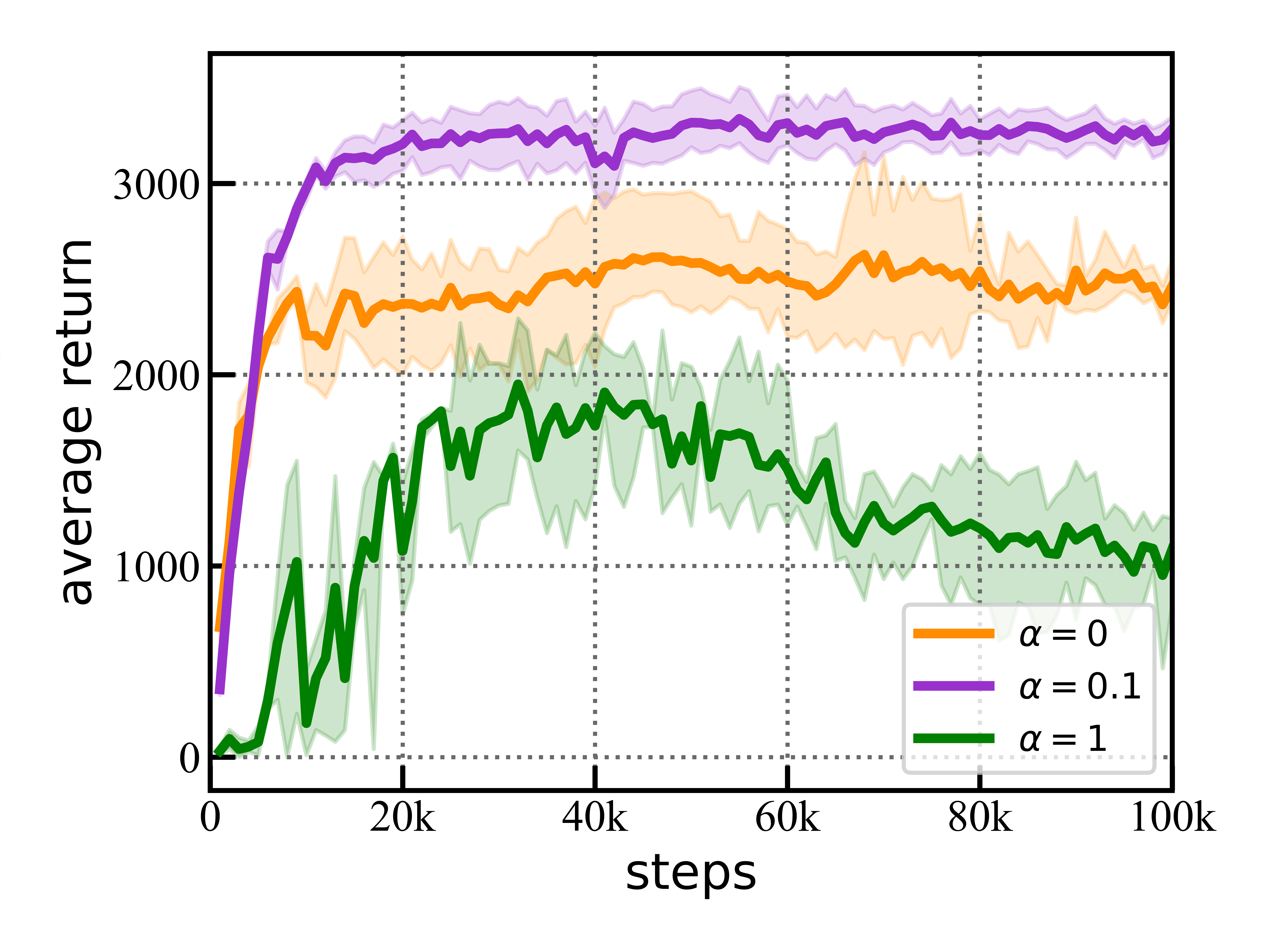}
}
\subfigure[Ablation on $\eta$.]{\includegraphics[width=4.05cm]{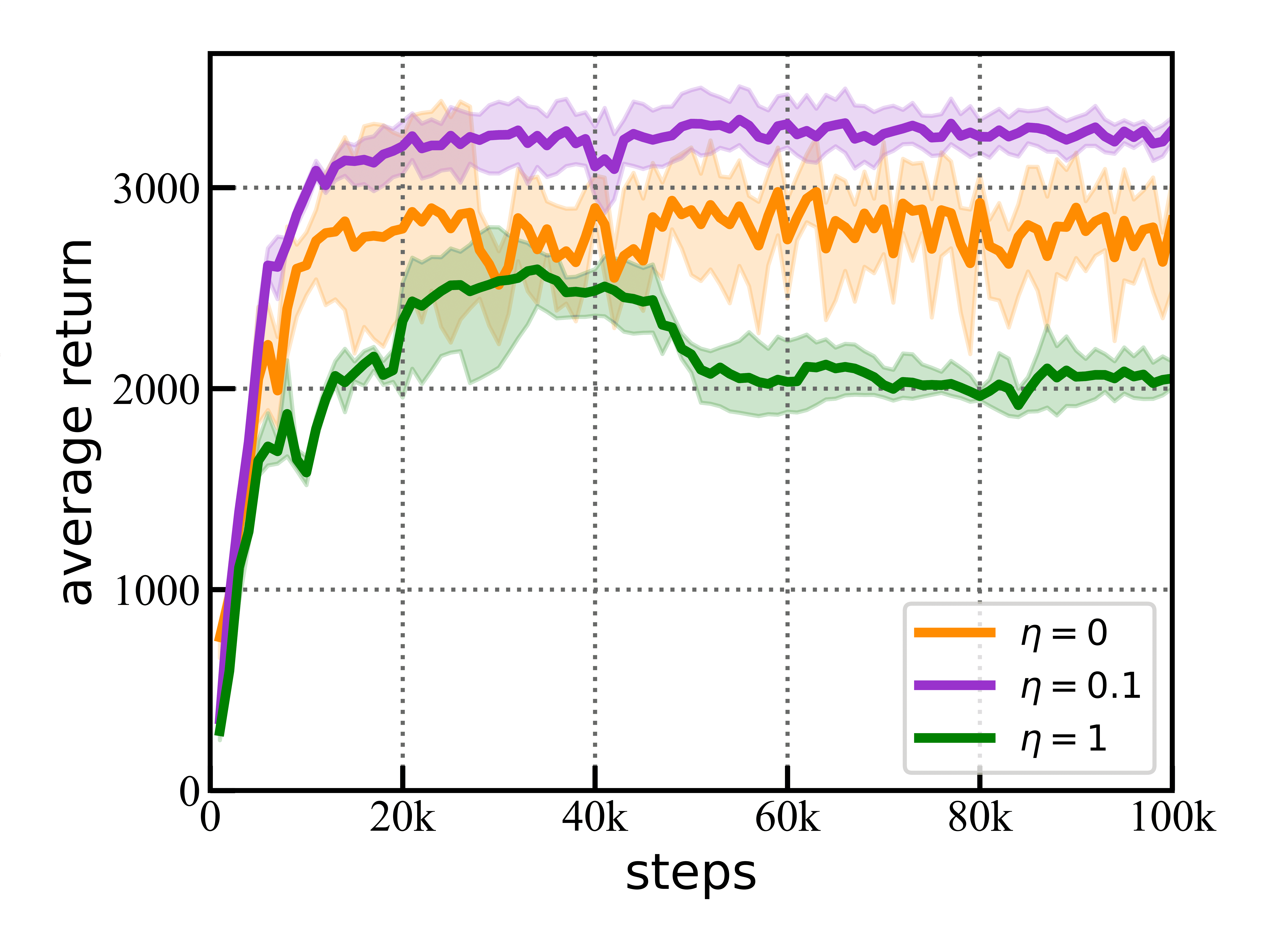}
}
\vspace{-2mm}
\caption{(a) Ablation study on adversarial learning hyperparameter $\alpha$. We set the hyperparameter to three different values, \{0, 0.1, 1\}, to examine its effect on the performance of the algorithm MOAN. (b) Ablation study on discrepancy measurement hyperparameter $\eta$.}
\label{fig:abla}
\end{center}
\vspace{-3mm}
\end{figure}

Figure~\ref{fig:abla}(a) shows the learning progress of our algorithm in the Hopper environment as a function of the hyperparameter $\alpha$. As shown, when $\alpha$ is too large, the generator (transition model) is heavily influenced by the noise produced by the discriminator, which can cause instability in the training process. On the other hand, when $\alpha$ is too small, the generator may not learn to generate diverse samples, resulting in plain performance. Therefore, it is important to choose an appropriate value for $\alpha$ to ensure stable and effective training of MOAN.

In addition, Figure~\ref{fig:abla}(b) demonstrates how MOAN's performance varies with different values of the hyperparameter $\eta$, which balances the importance of the reward value from the transition model and the deviation from the discriminator. When $\eta$ is too large, the reward may be too conservative for agent exploration in policy optimization. Therefore, it is important to choose an appropriate value for $\eta$ to balance exploration and exploitation in policy optimization.

In terms of sample complexity, the training of the discriminator has low sample complexity, since it only needs to distinguish between real and generated data. In our experiments, the discriminator consists of only two layers of networks with a simple structure, which makes it easy to train. Regarding the stability of convergence, the input state and action are usually low-dimensional vectors in the task of offline RL, making it easier to achieve stable and efficient training compared to high-dimensional image samples. As Figure \ref{fig:ab} shows, MOAN is stable under multiple random seeds. However, the value of hyperparameter $\alpha$ still affects the stability of the training process. Choosing an appropriate value for $\alpha$ is critical to achieving stable and effective training of MOAN.


Overall, selecting appropriate hyperparameters is crucial for MOAN's stability and performance in online environments. The results of this study show that incorporating adversarial learning and discrepancy penalty into MOAN can result in improved performance over prior algorithms. These findings have implications for the development of future reinforcement learning algorithms, particularly those designed for offline environments.

\section{Conclusion}
In this paper, we proposed an offline RL framework called MOAN, which introduces a two-player game to improve the generalization capability of the transition model and mitigate the negative effects of potentially problematic rollouts during offline reinforcement learning. Thus, MOAN can effectively balance the exploitation of logging dataset and exploration in out-of-distribution regions. Our experiments on several benchmark tasks demonstrated that our method achieved the highest performance in most settings. However, a major limitation of MOAN is the increased computational complexity required for the training of the discriminator network. In the future, we plan to develop more accurate and efficient strategies, such as improving the architecture of the discriminator network, to further enhance the performance and computational efficiency of MOAN. Additionally, we also plan to investigate the integration of various model-free approaches into MOAN to increase performance on certain datasets, such as the Walker2d \textit{medium-expert}.

\ack This paper is partially supported by National Natural Science Foundation of China under Grant No.62202238 and No.62276142; Natural Science Foundation of Jiangsu Province under Grant No.BK20200752.
\vspace{-4mm}
\bibliography{ecai}
\end{document}